\documentclass[11pt,a4paper]{article}

\usepackage[utf8]{inputenc}
\usepackage[T1]{fontenc}
\usepackage{times}
\usepackage{graphicx}
\usepackage{booktabs}
\usepackage{multirow}
\usepackage{amsmath}
\usepackage{hyperref}
\usepackage{xcolor}
\usepackage{tikz}
\usepackage{pgfplots}
\usepackage[margin=2.5cm]{geometry}
\usepackage{float}
\usepackage{natbib}
\usepackage{authblk}

\pgfplotsset{compat=1.18}
\usetikzlibrary{shapes, arrows.meta, positioning, calc, fit, patterns, decorations.pathreplacing}

\title{\textbf{SocialX: A Modular Platform for Multi-Source\\Big Data Research in Indonesia}}

\author[1]{Muhammad Apriandito Arya Saputra}
\author[2]{Andry Alamsyah}
\author[2]{Dian Puteri Ramadhani}
\author[3]{Thomhert Suprapto Siadari}
\author[4]{Hanif Fakhrurroja}

\affil[1]{SocialX \\ \texttt{apriandito@socialx.id}}
\affil[2]{Center of Excellence SAKTI, Research Institute Intelligent Business \& Sustainable Economy, Telkom University \\ \texttt{\{andry, dianpramadhani\}@telkomuniversity.ac.id}}
\affil[3]{Biomedical Engineering Study Program, School of Electrical Engineering, Telkom University \\ \texttt{thomhert@telkomuniversity.ac.id}}
\affil[4]{Research Center for Smart Mechatronics, National Research and Innovation Agency (BRIN) \\ \texttt{hani010@brin.go.id}}

\date{}

\begin{document}

\maketitle

\begin{abstract}
Big data research in Indonesia is constrained by a fundamental fragmentation: relevant data is scattered across social media, news portals, e-commerce platforms, review sites, and academic databases, each with different formats, access methods, and noise characteristics. Researchers must independently build collection pipelines, clean heterogeneous data, and assemble separate analysis tools, a process that often overshadows the research itself. We present SocialX, a modular platform for multi-source big data research that integrates heterogeneous data collection, language-aware preprocessing, and pluggable analysis into a unified, source-agnostic pipeline. The platform separates concerns across three independent layers (collection, preprocessing, and analysis) connected by a lightweight job coordination mechanism. This modularity allows each layer to grow independently: new data sources, preprocessing methods, or analysis tools can be added without modifying the existing pipeline. We describe the design principles that enable this extensibility, detail the preprocessing methodology that addresses challenges specific to Indonesian text across registers, and demonstrate the platform's utility through a walkthrough of a typical research workflow. SocialX is publicly accessible as a web-based platform at \url{https://www.socialx.id}.
\end{abstract}

\section{Introduction}

Indonesia produces an enormous volume of digital text. With over 185 million internet users \citep{datareportal2024}, data is generated continuously across social media platforms, news portals, e-commerce marketplaces, review sites, government portals, and academic repositories. For researchers studying public opinion, market dynamics, policy impact, or social phenomena, this data represents a rich and largely untapped resource.

The challenge is not the availability of data but the effort required to use it. Big data research typically demands three distinct capabilities: (1) collecting data from heterogeneous sources, each with its own structure, access protocol, and format; (2) transforming the collected data into a clean, research-ready dataset by handling noise, duplication, language mixing, and topical irrelevance; and (3) applying analytical methods such as sentiment analysis, network analysis, trend detection, and spatial analysis to extract meaningful patterns.

Each of these steps requires specialized engineering. Consider a researcher studying public perception of a new government policy. Relevant data might exist on Twitter (public reactions), Instagram (visual commentary), news portals (journalistic coverage and reader comments), e-commerce reviews (if the policy affects consumer products), and Google Maps reviews (if it affects public services or businesses). Each source has a different structure, a different access method, and a different type of noise. The researcher must build or configure separate scrapers, reconcile different data formats, filter out irrelevant content in multiple registers of Indonesian, and then apply one or more analysis methods, all before the actual research can begin.

This fragmentation creates a high barrier to entry. It privileges researchers with engineering resources and limits the scope of studies to whatever sources a given team can practically handle. Many studies default to a single data source, typically Twitter, not because it best answers the research question, but because it is the most accessible.

Existing solutions address parts of this problem. Commercial platforms such as Brandwatch and Meltwater provide integrated pipelines but are expensive, limited to social media, and not optimized for Bahasa Indonesia \citep{wilie2020indonlu}. Open-source frameworks such as GATE \citep{cunningham2002gate} and Orange \citep{demsar2013orange} offer analysis capabilities but assume data has already been collected and cleaned. Indonesian NLP models exist \citep{wilie2020indonlu} but as standalone components, not as part of an integrated research workflow. No existing tool provides a unified pipeline from heterogeneous data collection through language-aware preprocessing to multi-method analysis for Indonesian big data research.

We present SocialX, a web-based platform designed to close this gap. SocialX integrates multi-source data collection, language-aware preprocessing, and pluggable analysis into a single platform accessible from the browser. The platform is built on three design principles:

\begin{enumerate}
    \item \textbf{Source agnosticism.} The platform treats all data sources (social media, news, e-commerce, reviews, academic databases) through a uniform interface. Collection connectors normalize heterogeneous data into a common schema, making downstream processing independent of data origin.

    \item \textbf{Modularity.} The platform separates collection, preprocessing, and analysis into independent layers that can grow without affecting each other. Adding a new data source requires no changes to the preprocessing or analysis layers, and vice versa.

    \item \textbf{Language awareness.} Indonesian text spans a wide register spectrum, from formal news prose to highly informal social media slang with code-mixing. The preprocessing layer is designed to handle this spectrum, including a context-conditioned relevancy classifier trained specifically for Indonesian \citep{saputra2026indobert}.
\end{enumerate}

This paper describes the architecture and methodology behind SocialX. Rather than cataloging individual features, we focus on the design decisions that make the platform extensible and the preprocessing methodology that addresses challenges specific to Indonesian big data research.

\section{Related Work}

\subsection{Commercial Platforms}

Tools such as Brandwatch, Meltwater, and Sprinklr offer end-to-end social media monitoring with data collection, filtering, and sentiment analysis. These platforms are designed primarily for brand monitoring and market research, with pricing models typically prohibitive for academic use. They focus almost exclusively on social media, offering limited or no integration with news portals, e-commerce data, review platforms, or academic sources. Their NLP pipelines are optimized for high-resource languages; Indonesian support, when available, handles formal text adequately but struggles with informal registers \citep{wilie2020indonlu}.

\subsection{Open-Source Frameworks}

GATE \citep{cunningham2002gate} provides a mature plugin-based architecture for text processing that has influenced our design. However, GATE focuses on text annotation and does not include data collection capabilities. Orange \citep{demsar2013orange} offers visual data analysis with text mining extensions but similarly assumes data has been pre-collected. Neither provides Indonesian-specific NLP components or multi-source data collection.

Platform-specific scraping tools (e.g., snscrape, Twint) provide access to individual data sources but require programming knowledge, produce source-specific output formats, and include no analysis capabilities. Assembling these tools into a coherent research pipeline remains the researcher's responsibility.

\subsection{Indonesian NLP}

IndoNLU \citep{wilie2020indonlu} established benchmarks for Indonesian NLP, and pre-trained models such as IndoBERT provide foundations for downstream tasks including sentiment analysis \citep{mdhugol2022sentiment}, named entity recognition, and natural language inference. These models are released as standalone components. Integrating them into a research workflow (loading the model, preparing inputs, interpreting outputs, handling batch processing) still requires significant engineering for each project.

\subsection{The Integration Gap}

The common limitation across these categories is fragmentation. Commercial tools cover collection and analysis but are source-limited and language-constrained. Open-source frameworks cover analysis but not collection. NLP models cover specific tasks but not workflows. Table~\ref{tab:comparison} summarizes this landscape.

\begin{table}[H]
\centering
\caption{Comparison of existing approaches for big data research. SocialX integrates multi-source collection, language-aware preprocessing, and pluggable analysis with Indonesian language support.}
\label{tab:comparison}
\small
\begin{tabular}{lccccc}
\toprule
\textbf{Approach} & \textbf{Multi-Source} & \textbf{Preprocessing} & \textbf{Analysis} & \textbf{Indonesian} & \textbf{No-Code} \\
 & \textbf{Collection} & & & \textbf{NLP} & \\
\midrule
Brandwatch / Meltwater & Social only & Keyword & Sentiment & Limited & Yes \\
GATE & No & Yes & Yes & No & No \\
Orange & No & Limited & Yes & No & Yes \\
snscrape / Twint & Single source & No & No & No & No \\
IndoBERT (standalone) & No & No & Single task & Yes & No \\
\midrule
\textbf{SocialX} & \textbf{Yes} & \textbf{Yes (ML)} & \textbf{Yes} & \textbf{Yes} & \textbf{Yes} \\
\bottomrule
\end{tabular}
\end{table}

\section{Platform Architecture}

\begin{figure}[H]
\centering
\begin{tikzpicture}[
    layer/.style={rectangle, draw=black, thick, rounded corners=3pt,
                  minimum width=11.5cm, minimum height=1.8cm, fill=white},
    module/.style={rectangle, draw=black!50, thick, rounded corners=2pt,
                   minimum width=2.3cm, minimum height=0.7cm,
                   fill=black!3, font=\small, align=center},
    store/.style={rectangle, draw=black, thick, rounded corners=3pt,
                  minimum width=5cm, minimum height=0.7cm, fill=black!6,
                  font=\small\bfseries},
    ltitle/.style={font=\small\bfseries, anchor=north west},
    arr/.style={-{Stealth[length=3mm]}, very thick},
    darr/.style={-{Stealth[length=2.5mm]}, thick, dashed, black!40},
    note/.style={font=\scriptsize\itshape, text=black!50},
]


\node[layer] (CL) at (0, 0) {};
\node[ltitle] at ([shift={(4pt,-2pt)}]CL.north west) {Collection Layer};
\node[module] at (-3.4, -0.25) {Social Media};
\node[module] at (-0.8, -0.25) {News Portals};
\node[module] at (1.8, -0.25) {Reviews};
\node[module] at (4.2, -0.25) {Academic};

\node[store] (DS) at (0, -1.6) {Unified Data Store};
\node[note, right=0.3cm of DS] {common schema};

\node[layer] (PL) at (0, -3.2) {};
\node[ltitle] at ([shift={(4pt,-2pt)}]PL.north west) {Preprocessing Layer};
\node[module] at (-3.4, -3.45) {Deduplication};
\node[module] at (-0.8, -3.45) {Language Det.};
\node[module] at (1.8, -3.45) {Keyword Filter};
\node[module] at (4.2, -3.45) {Relevancy};

\node[layer] (AL) at (0, -5.4) {};
\node[ltitle] at ([shift={(4pt,-2pt)}]AL.north west) {Analysis Layer};
\node[module] at (-3.4, -5.65) {Sentiment};
\node[module] at (-0.8, -5.65) {Network};
\node[module] at (1.8, -5.65) {Trend};
\node[module] at (4.2, -5.65) {Spatial};

\draw[arr] ([yshift=-2pt]CL.south) -- ([yshift=2pt]DS.north);
\draw[arr] ([yshift=-2pt]DS.south) -- ([yshift=2pt]PL.north);
\draw[arr] ([yshift=-2pt]PL.south) -- ([yshift=2pt]AL.north);

\node[rectangle, draw=black!50, thick, dashed, rounded corners=3pt,
      minimum width=2cm, minimum height=5.2cm, fill=white] (JC) at (7.8, -2.7) {};
\node[font=\small\bfseries, align=center] at (7.8, -1) {Job\\[-2pt]Coordinator};
\node[font=\scriptsize, align=center, text=black!60] at (7.8, -2.5) {create\\execute\\complete};
\node[note] at (7.8, -4) {database-driven};

\draw[darr] ([xshift=-2pt]JC.west |- CL.east) -- (CL.east);
\draw[darr] ([xshift=-2pt]JC.west |- PL.east) -- (PL.east);
\draw[darr] ([xshift=-2pt]JC.west |- AL.east) -- (AL.east);

\end{tikzpicture}
\caption{SocialX platform architecture. Three independent layers are connected through a unified data store and coordinated by a database-driven job queue. Each layer is independently extensible.}
\label{fig:architecture}
\end{figure}
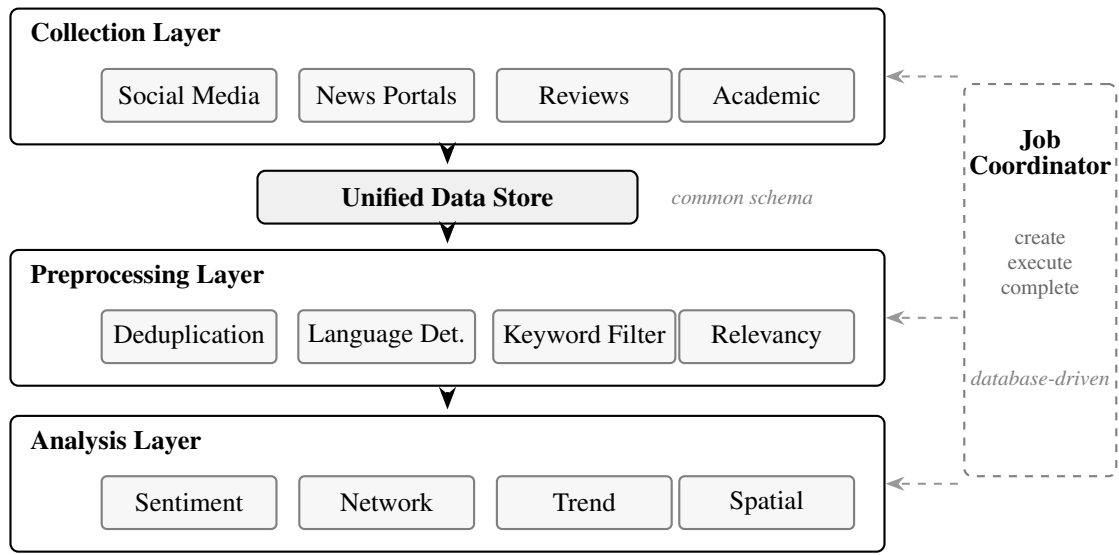

SocialX is organized into three layers (collection, preprocessing, and analysis) connected by a unified data store and a lightweight job coordination mechanism. Figure~\ref{fig:architecture} illustrates this architecture.

\subsection{Design Principle: Independence Through Simplicity}

The central architectural decision in SocialX is the separation of layers through a shared data store rather than direct inter-layer communication. When a user initiates a task (collecting data, preprocessing a dataset, or running an analysis), the system creates a job record in the database. Independent worker processes continuously poll for pending jobs matching their capability, execute them, store the results, and update the job status.

This design enforces a clean separation: each worker knows only how to execute its own task type and how to read from and write to the data store. A collection worker does not know that preprocessing exists; an analyzer does not know how data was collected or from which source. This is what makes the platform genuinely source-agnostic. A dataset that originated from Twitter is indistinguishable from one that originated from news articles or e-commerce reviews by the time it reaches the analysis layer.

The simplicity of this coordination mechanism is deliberate. It requires no infrastructure beyond the database that already stores the data, making deployment and maintenance straightforward. More importantly, it means that adding a new component to any layer requires implementing only a single interface: produce or consume data in the common schema.

\subsection{Collection Layer}

The collection layer provides connectors to heterogeneous data sources organized into four categories:

\begin{itemize}
    \item \textbf{Social media}: platforms where users generate short-form content (posts, comments, replies).
    \item \textbf{News portals}: editorial content from major Indonesian news outlets, including article text and reader comments.
    \item \textbf{E-commerce and reviews}: product listings, user reviews, and ratings from marketplaces and review platforms.
    \item \textbf{Academic sources}: research papers and citation data from scholarly databases.
\end{itemize}

Each connector normalizes its source-specific output into the platform's common schema. This normalization is the key abstraction: regardless of whether data comes from a social media API, a news website, or a review platform, it enters the data store in the same format. Downstream layers operate on this normalized data without knowledge of its origin. The set of supported sources continues to grow; importantly, none of the additions to date have required changes to the preprocessing or analysis layers.

\subsection{Preprocessing Layer}

The preprocessing layer transforms raw collected data into research-ready datasets through composable filters. Because it operates on the normalized schema rather than source-specific formats, the same preprocessing pipeline works identically on data from any source. This layer is described in detail in Section~\ref{sec:methodology}, as it contains the platform's primary methodological contribution.

\subsection{Analysis Layer}

The analysis layer provides a growing set of analytical tools, each implemented as an independent module. Current capabilities span multiple analytical paradigms: transformer-based text classification (sentiment, emotion, zero-shot categorization), network analysis (social networks, entity relationships, co-authorship), temporal analysis (trend detection, time-series patterns), and spatial analysis (geographic distribution, competitor mapping). All analyzers follow a uniform interface, accepting a dataset reference and configuration parameters and producing structured outputs rendered as interactive visualizations.

Each analyzer operates on a single preprocessed dataset at a time. The diversity of available analysis types reflects the platform's multi-source nature: social network analysis is most applicable to social media data, spatial analysis applies to location-based data, and co-authorship networks apply to academic data.

\section{Methodology: Language-Aware Preprocessing}
\label{sec:methodology}

The preprocessing layer addresses the most significant practical challenge in Indonesian big data research: transforming noisy, multilingual, and often off-topic data from heterogeneous sources into a clean dataset that faithfully represents the research topic.

\subsection{The Noise Problem Across Sources}

Raw data collected from any online source is inherently noisy, but the nature of noise varies by source type. Social media data contains slang, code-mixing, and off-topic posts captured by ambiguous keywords. News data contains boilerplate text, advertisements, and articles that mention a keyword in passing without being substantively about the topic. E-commerce reviews contain spam, template reviews, and content in unexpected languages. Academic data may include papers that cite a topic peripherally without contributing to it.

Despite these differences, the fundamental preprocessing operations (deduplication, language identification, topical filtering) apply across all source types. SocialX's preprocessing layer exploits this commonality by operating on the normalized schema, applying the same filter pipeline regardless of data origin.

\subsection{Composable Filter Pipeline}

Preprocessing in SocialX is implemented as a pipeline of composable filters that researchers configure through the web interface. The five filters are applied sequentially:

\begin{enumerate}
    \item \textbf{Deduplication.} Removes exact and near-duplicate entries based on content hash or URL. When merging data from multiple sources, duplication is common: the same news story may appear across multiple portals, and the same opinion may be cross-posted across platforms.

    \item \textbf{Date filtering.} Restricts the dataset to a specific time window, essential for event-driven research where temporal boundaries define the study scope.

    \item \textbf{Language detection.} Identifies the language of each text and filters to the target language(s). Indonesian online data contains significant amounts of English, regional languages (Javanese, Sundanese), and code-mixed text.

    \item \textbf{Keyword filtering.} Applies inclusion and exclusion keyword lists for coarse-grained topical filtering. Effective for removing obviously unrelated content but insufficient for nuanced topical relevance.

    \item \textbf{Context-conditioned relevancy classification.} A fine-tuned IndoBERT model \citep{saputra2026indobert} that evaluates whether each text is relevant to a user-specified research context. Unlike keyword filtering, this model captures implicit relevance: texts that discuss a topic through personal experiences, indirect references, or informal language without using explicit keywords.
\end{enumerate}

Researchers can enable, disable, and configure each filter independently. This composability is important because different research questions require different preprocessing strategies. A study with precise temporal boundaries may emphasize date filtering; a study on a diffuse social phenomenon may rely heavily on the relevancy classifier; a multilingual study may skip language filtering entirely.

\subsection{Context-Conditioned Relevancy Classification}

The relevancy classifier deserves particular attention as the most methodologically distinctive component. The classifier is based on IndoBERT Large \citep{wilie2020indonlu}, fine-tuned on 31,360 labeled text-context pairs across 188 topics \citep{saputra2026indobert}. Given a research context (e.g., ``fuel price policy and its impact on daily life'') and a candidate text from any source, the model produces a binary relevancy judgment. The input is formatted as a sentence pair:

\begin{center}
\texttt{[CLS] context [SEP] text [SEP]}
\end{center}

The model achieves an F1 score of 0.948 across diverse text registers, from formal news prose to highly informal social media. Its training data was constructed through an iterative, failure-driven process that specifically targeted weaknesses on informal and implicit text \citep{saputra2026indobert}. In the SocialX pipeline, relevancy classification serves as the final and most semantically rich filter, applied after coarser filters have already removed obvious noise.

Table~\ref{tab:preprocessing-impact} provides the detailed breakdown of the preprocessing pipeline's cumulative effect.

\begin{table}[H]
\centering
\caption{Cumulative effect of preprocessing filters on a multi-source dataset collected for the topic ``public response to fuel price increases.'' Data was collected from social media platforms and news portals.}
\label{tab:preprocessing-impact}
\small
\begin{tabular}{lrrr}
\toprule
\textbf{Stage} & \textbf{Records} & \textbf{Removed} & \textbf{Reduction} \\
\midrule
Raw collected data & 12,847 & & \\
After deduplication & 10,203 & 2,644 & $-$20.6\% \\
After language detection & 9,451 & 752 & $-$7.4\% \\
After keyword filtering & 7,832 & 1,619 & $-$17.1\% \\
After relevancy classification & 5,614 & 2,218 & $-$28.3\% \\
\midrule
\textbf{Total} & \textbf{5,614} & \textbf{7,233} & \textbf{$-$56.3\%} \\
\bottomrule
\end{tabular}
\end{table}

The relevancy classifier alone removed 28.3\% of texts that survived all prior filters. Manual inspection of a random sample of 200 removed texts confirmed that over 90\% were genuinely off-topic, validating the classifier's contribution to data quality across source types.

\section{Platform Walkthrough}
\label{sec:walkthrough}

To illustrate how SocialX works in practice, we walk through a typical research workflow: a researcher investigating public discourse around fuel price policy in Indonesia. Table~\ref{tab:walkthrough} summarizes the four steps.

\begin{table}[H]
\centering
\caption{Research workflow in SocialX. Each step is handled through the web interface.}
\label{tab:walkthrough}
\small
\begin{tabular}{clp{7.5cm}}
\toprule
\textbf{Step} & \textbf{Action} & \textbf{Description} \\
\midrule
1 & Select sources & Choose data sources from catalog. Configure keywords, date range, and source-specific parameters. \\
2 & Collect data & Platform runs scrapers automatically. Output is normalized to the common schema and stored. \\
3 & Preprocess & Select collected datasets as input. Enable and configure filters: deduplication, language detection, keyword, relevancy. \\
4 & Analyze & Choose an analyzer (sentiment, network, trend, etc.). Results are rendered as interactive visualizations in the browser. \\
\bottomrule
\end{tabular}
\end{table}

\subsection{Step 1: Source Selection and Configuration}

The researcher begins by selecting data sources from the platform's collection catalog. For this study, she selects Twitter (with keywords ``BBM,'' ``bensin,'' ``Pertamina''), two Indonesian news portals, and Instagram (with hashtags related to fuel prices). Each source is configured independently through the web interface: keywords, date ranges, and source-specific parameters such as hashtags or search queries. This replaces the manual process of finding, installing, and configuring separate scraping tools for each platform.

\subsection{Step 2: Data Collection}

After configuration, the researcher initiates collection. The platform's worker processes execute each scraper automatically, normalize the output to the common schema, and store the results. The researcher can monitor progress through the interface and see how many records have been collected from each source. Each source produces data in a different raw format, but the researcher never interacts with these formats directly.

\subsection{Step 3: Preprocessing}

With the raw data collected, the researcher creates a preprocessing job. She selects the collected datasets as input, then configures the filter pipeline:

\begin{itemize}
    \item \textbf{Deduplication}: enabled, to remove cross-posted content
    \item \textbf{Language detection}: filter to Indonesian only
    \item \textbf{Keyword filtering}: exclude posts containing ``BlackBerry Messenger'' (to disambiguate ``BBM'')
    \item \textbf{Relevancy classification}: enabled, with context ``Kebijakan harga BBM dan dampaknya terhadap kehidupan sehari-hari'' (Fuel pricing policy and its impact on daily life)
\end{itemize}

The platform runs the filter pipeline and produces a cleaned dataset. In a representative run, preprocessing reduced the dataset by over 50\%, with the relevancy classifier removing the largest single proportion of noise (Table~\ref{tab:preprocessing-impact}).

\subsection{Step 4: Analysis}

The researcher selects an analyzer from the platform's catalog, chooses sentiment analysis, specifies the preprocessed dataset and text column, and submits the job. The platform runs the analysis and renders the results as interactive visualizations in the browser. She can then run additional analyzers (trend detection, network analysis, word cloud) on the same dataset without writing code or managing model dependencies.

\subsection{What This Replaces}

Table~\ref{tab:workflow-comparison} contrasts this workflow with the manual alternative.

\begin{table}[H]
\centering
\caption{Manual workflow versus SocialX for a multi-source research project.}
\label{tab:workflow-comparison}
\small
\begin{tabular}{lp{4.5cm}p{4.5cm}}
\toprule
\textbf{Step} & \textbf{Manual Approach} & \textbf{SocialX} \\
\midrule
Collection & Install and configure separate scraping tools per platform. Handle authentication, rate limits, and errors in code. & Select sources and configure keywords through web interface. \\
\addlinespace
Format reconciliation & Write scripts to parse and unify different output formats into a single structure. & Handled automatically by collection connectors. \\
\addlinespace
Preprocessing & Write cleaning scripts. Install and run NLP models (language detection, relevancy) locally. & Configure filters through web interface. ML models run on backend. \\
\addlinespace
Analysis & Load pre-trained models, write inference code, build visualizations. & Select analyzer, submit job, view results in browser. \\
\addlinespace
Requirements & Python, NLP libraries, GPU (optional), programming skills. & Web browser. \\
\bottomrule
\end{tabular}
\end{table}

\section{Extensibility and Growth}

A research platform must accommodate change. Data sources appear and disappear, new analytical methods emerge, and research questions evolve. SocialX's modular architecture is designed for this.

Since initial deployment, the collection layer has grown substantially in the number of supported sources, and the analysis layer has expanded to cover multiple analytical paradigms. The preprocessing layer has evolved from simple keyword filtering to a multi-stage pipeline incorporating machine learning. None of these changes required restructuring the platform. Each new component was added by implementing the relevant interface and registering a worker.

The relevancy classifier illustrates this well. It was developed and published as a standalone model \citep{saputra2026indobert} and later integrated into the preprocessing layer by writing an adapter. The collection and analysis layers were unaware of the change. This pattern, developing capabilities independently and integrating them through the common interface, is how we expect the platform to continue growing.

The cost of adding new components is deliberately low: a new data source requires a connector that outputs the common schema; a new filter requires a function that transforms a dataset; a new analyzer requires a module that produces structured output. In each case, the new component interacts only with the data store, not with other components.

\section{Discussion}

\subsection{Reducing the Engineering Overhead}

Our experience building and operating SocialX confirms what many researchers know implicitly: most of the time spent on big data research goes into data engineering rather than data analysis. Collection, format reconciliation, cleaning, and filtering are necessary steps, but they do not directly advance the research question. SocialX automates these steps behind a web interface, allowing researchers to spend their time on study design and interpretation instead.

This matters especially in the Indonesian academic context, where research teams vary widely in their access to engineering support. A platform that requires only a browser removes a significant practical barrier.

\subsection{Preprocessing as a Methodological Choice}

The preprocessing layer is not just a technical convenience. Which data to include, which to exclude, and by what criteria are methodological decisions that shape research outcomes. SocialX makes these decisions explicit, configurable, and reproducible through the filter pipeline. The relevancy classifier, in particular, enables a form of data curation that goes beyond keyword matching: specifying a research context and applying relevancy classification amounts to a semantic judgment about topical inclusion, applied consistently at scale.

\subsection{Current Limitations and Future Directions}

The platform has several limitations worth noting. The analysis layer currently operates on one dataset at a time; automated cross-source comparison (e.g., contrasting sentiment across social media and news for the same topic) is not yet supported and is a clear next step. The relevancy classifier was trained on 188 topics and may underperform on domains not represented in its training data. The platform processes data in batch mode, with no real-time streaming capability. And while the architecture is source-agnostic, the current set of connectors is focused on Indonesian platforms.

On the ethical side, SocialX collects only publicly available data and does not circumvent access controls. Researchers using the platform are responsible for complying with applicable regulations and institutional ethics requirements, particularly regarding privacy when combining data from multiple sources.

\section{Conclusion}

SocialX is a modular platform that brings together multi-source data collection, language-aware preprocessing, and pluggable analysis for big data research in Indonesia. Its architecture separates these concerns into independent layers connected through a common data schema, an approach that has proven extensible in practice as sources, filters, and analyzers have been added over time without requiring structural changes.

The preprocessing methodology, built around composable filters and a context-conditioned relevancy classifier for Indonesian text, addresses a gap that affects the quality of all downstream work: the transformation of noisy, heterogeneous web data into research-ready datasets. The platform is accessible at \url{https://www.socialx.id}.

\bibliographystyle{plainnat}

\end{document}